\documentclass{article}
\usepackage{spconf,amsmath,graphicx,hyperref}
\usepackage{marvosym}
\usepackage{amsmath}
\usepackage{amssymb}
\usepackage{booktabs}   
\usepackage{multirow}    
\usepackage{makecell}    
\usepackage{adjustbox}

\title{RACap: Relation-Aware Prompting \\ for Lightweight Retrieval-Augmented Image Captioning}
\name{    Xiaosheng Long\textsuperscript{\rm 1},
    Hanyu Wang\textsuperscript{\rm 1},
    Zhentao Song\textsuperscript{\rm 1},
    Kun Luo$^{*}$\textsuperscript{\rm 2}\thanks{* : Corresponding Author (luokun2822@sohu.com)} and Hongde Liu$^{**}$\textsuperscript{\rm 1}\thanks{** : Principal Corresponding Author (liuhongde@seu.edu.cn)}
}
\address{
\textsuperscript{\rm 1}School of biological science and medical engineering, Southeast University, China\\
\textsuperscript{\rm 2}Department of Neurosurgery, The second affiliated hospital of Xinjiang Medical University, China
}

\begin{document}
\ninept
\maketitle
\begin{abstract}
Recent retrieval-augmented image captioning methods incorporate external knowledge to compensate for the limitations in comprehending complex scenes. However, current approaches face challenges in relation modeling: (1) the representation of semantic prompts is too coarse-grained to capture fine-grained relationships; (2) these methods lack explicit modeling of image objects and their semantic relationships. To address these limitations, we propose RACap, a relation-aware retrieval-augmented model for image captioning, which not only mines structured relation semantics from retrieval captions, but also identifies heterogeneous objects from the image. RACap effectively retrieves structured relation features that contain heterogeneous visual information to enhance the semantic consistency and relational expressiveness. Experimental results show that RACap, with only 10.8M trainable parameters, achieves superior performance compared to previous lightweight captioning models.
\end{abstract}
\begin{keywords}
Image captioning, Lightweight vision-language models, Retrieval-augmented generation, Cross-Modal
\end{keywords}

\section{Introduction}
\label{sec:intro}

Approaches based on large language models (LLMs) have shown promising results in image captioning tasks \cite{45_li2020oscar, 17_li2022blip}. These methods primarily focus on training large-scale models with ever-growing datasets, which brings significant computational and resource burdens. 
Some models integrate pre-trained vision encoders with LLMs via a lightweight projection module to bridge visual and textual representations, such as BLIP-2 \cite{5_li2023blip}, ClipCap \cite{7_mokady2021clipcap}, and I-Tuning \cite{8_luo2023tuning}. Although this strategy significantly reduces trainable parameters, it often requires fine-tuning on novel data in order to achieve great performance.

To address this issue, recent models such as SmallCap \cite{9_ramos2023smallcap}, EVCap \cite{11_li2024evcap}, and ViPCap \cite{12_kim2025vipcap} adopt Retrieval-Augmented Generation (RAG) to incorporate external knowledge. As shown in Figure 1, this strategy not only reduces computational costs but also improves the quality of captions. These methods enhance the ability to describe images, but they overlook the spatial and semantic relationships among fine-grained objects.

In this paper, we present RACap, a relation-aware prompting for lightweight image captioning, which focuses on identifying object relationships accurately. Specifically, RACap extracts Subject–Predicate–Object–Environment (S-P-O-E) tuples from retrieval captions to uncover relations from the textual information. Compared to the original retrieval captions, this structured information contains less noise and better reflects object relationships. In addition, we introduce an object-aware module that is based on the slot-attention mechanism \cite{13_locatello2020object} to identify heterogeneous visual entities and extract key visual elements from the image. A slot retrieval module is proposed to retrieve the most relevant textual features for each heterogeneous visual slot to align semantic information with individual objects. These relational features are then injected into the image feature so that RACap can fully capture object relationships. With only 10.8M trainable parameters and 222.7M total parameters, RACap outperforms numerous baseline models on the COCO \cite{14_lin2014microsoft}, Flickr30k \cite{15_plummer2015flickr30k}, and NoCaps \cite{16_agrawal2019nocaps} datasets and presents strong adaptability to out-of-domain data without further fine-tuning or re-training.

\begin{figure}[t]
    \centering
    \hspace*{-0.6cm} 
    \includegraphics[width=6.2cm]{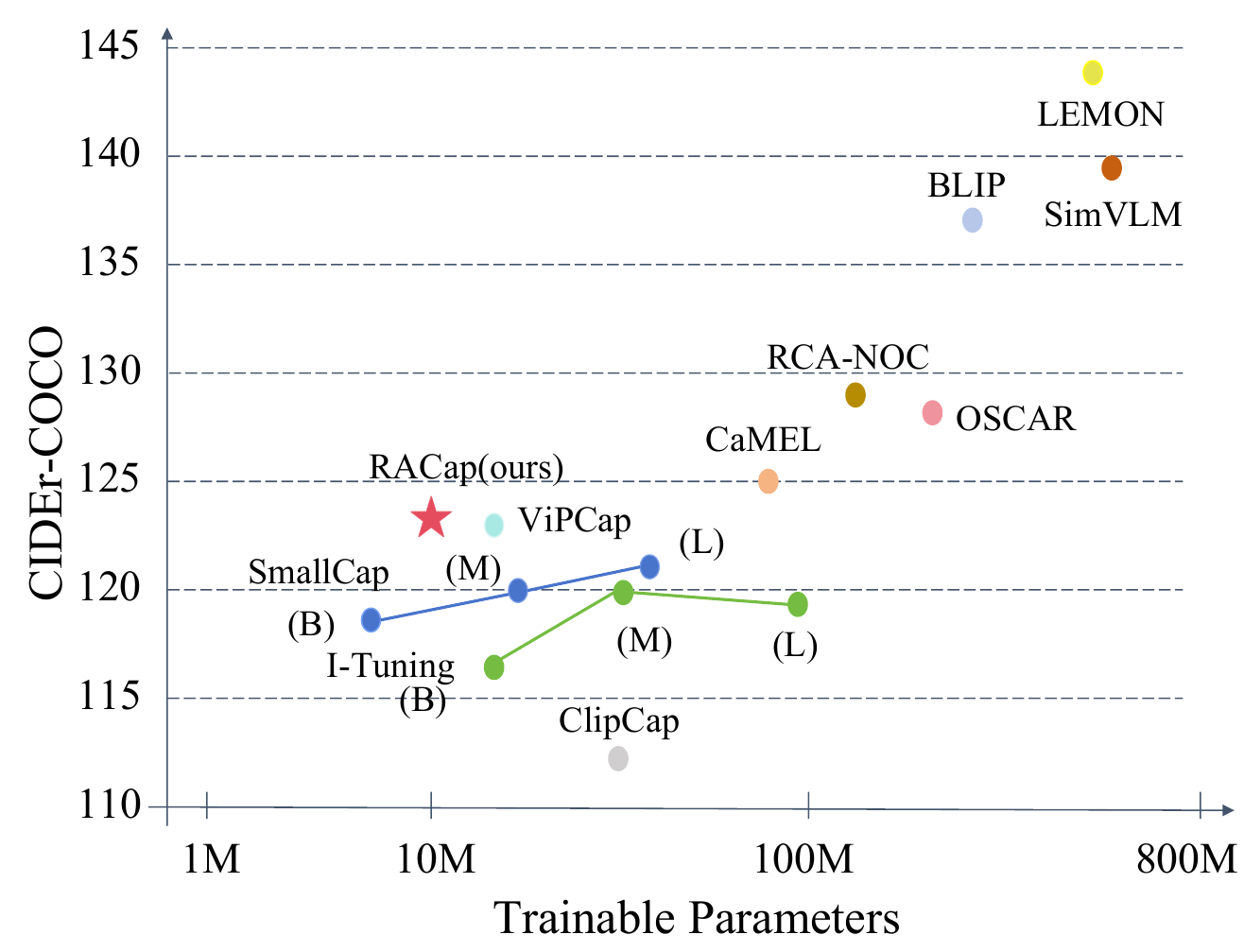}
    \vspace{-0.4cm}
    \caption{Comparison of trainable parameters and CIDEr scores. B, M, and L denote the Base, Medium, and Large models.}
    \label{fig1}
\end{figure}

\vspace{-0.3cm}
\section{Related Works}
\vspace{-0.2cm}
\subsection{Image Captioning.}
\vspace{-0.1cm}
Image captioning aims to generate descriptive texts given an input image. Large-scale vision-and-language (V\&L) models typically require substantial computational resources for training and deployment. To alleviate this issue, recent works have adopted pre-trained models and frozen most parameters, training only a small portion of the network to reduce resource consumption \cite{5_li2023blip,8_luo2023tuning, 46_fei2023transferable}. These approaches commonly freeze both the vision encoder and the language decoder, then train a lightweight projector module to align visual and textual representations. In RACap, we also adopt pre-trained models. The CLIP encoder is used to extract image and text features. GPT-2 is used as a text decoder to generate image captions.

\begin{figure*}[t]
  \centering
  \includegraphics[width=0.85\textwidth]{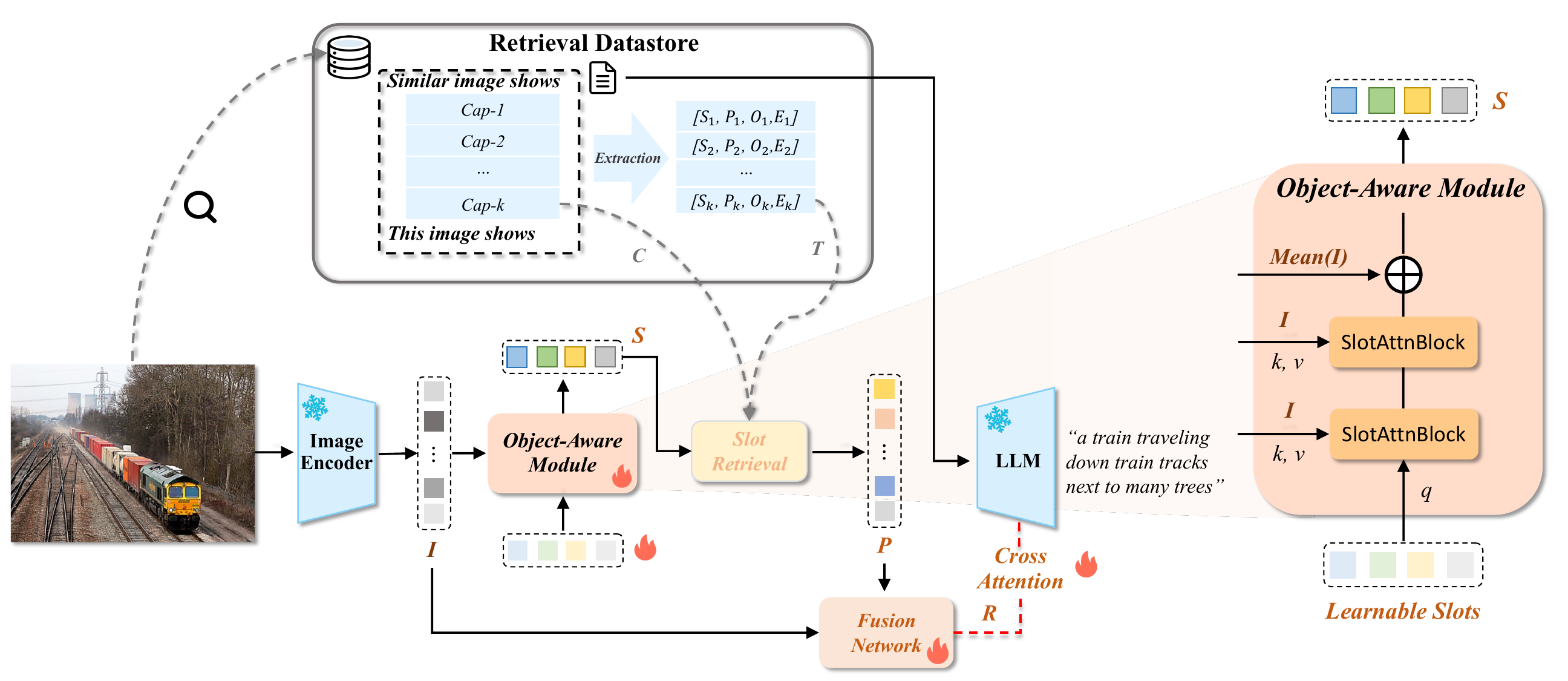}
  \vspace{-0.4cm} 
  \caption{An overview of RACap framework for Lightweight Image Captioning. The object-aware module learns heterogeneous slots \textit{S} from image patch features \textit{I}. Then the slot retrieval module retrieves the most correlated vectors from textual features \textit{C} (captions) and \textit{T} (S-P-O-E tuples extracted from retrieval datastore) based on similarity with \textit{S} to form the relation representation \textit{P}. Finally, the Fusion Network fuses \textit{P} into \textit{I} to produce a relation-aware visual prompt \textit{R}, which is passed to the text decoder via the cross-attention layer.}
\end{figure*}

\subsection{Retrieval-Augmented Generation}
Incorporating external memory has been shown to be a resource-efficient approach for image captioning by utilizing additional information retrieved from external memory. SmallCap \cite{9_ramos2023smallcap} leverages retrieval captions as prompts for the language generation model. MeaCap \cite{27_zeng2024meacap} employ a merge-and-filter strategy to process retrieval captions. Different from these approaches, RACap not only leverages retrieved captions relevant to the input image, but also extracts Subject–Predicate–Object–Environment (S-P-O-E) tuples from captions to uncover relation information in the text.
\vspace{-0.3cm}
\subsection{Prompting Program}
Prompting is a common technique for conveying instructions to language models. Recent works have also explored learnable embeddings as prompts. ViPCap \cite{12_kim2025vipcap} introduces a visual prompting approach guided by patch features for retrieved captions. EVCAP \cite{11_li2024evcap} combines learned visual features with object names to form an image prompt for open-world understanding. In contrast to previous methods, we propose a visual prompting approach that fully exploits both the retrieved captions and the visual features of the objects.

\vspace{-0.2cm}
\section{Proposed Approach}
\vspace{-0.1cm}
\subsection{Overall Structure}
RACap, as illustrated in Figure 2, aims to enhance lightweight captioning performance by integrating structured relation representations with heterogeneous visual information. Our approach is built on pre-trained models, namely CLIP and GPT-2, and leverages a retrieval datastore.
RACap parses the retrieval captions from retrieval datastore into S-P-O-E tuples which are encoded to \textit{T} (from S-P-O-E tuples) and \textit{C} (from retrieval captions). What's more, RACap initializes a set of learnable slots \textit{S} to interact with the image patch features \textit{I} via the object-aware module to iteratively capture the heterogeneous semantics in the image. The slot retrieval module then retrieves the highly relevant vectors from \textit{C} and \textit{T} with respect to \textit{S}, and concatenates them to form the relational features \textit{P}. Finally, \textit{P} and \textit{I} are fed into the Fusion Network, resulting in a relation-aware image feature \textit{R}. The cross-attention layer in GPT-2 is trained to integrate \textit{R} into the decoding process for caption generation.

\begin{figure}[t]
  \centering
  \includegraphics[width=0.4\textwidth]{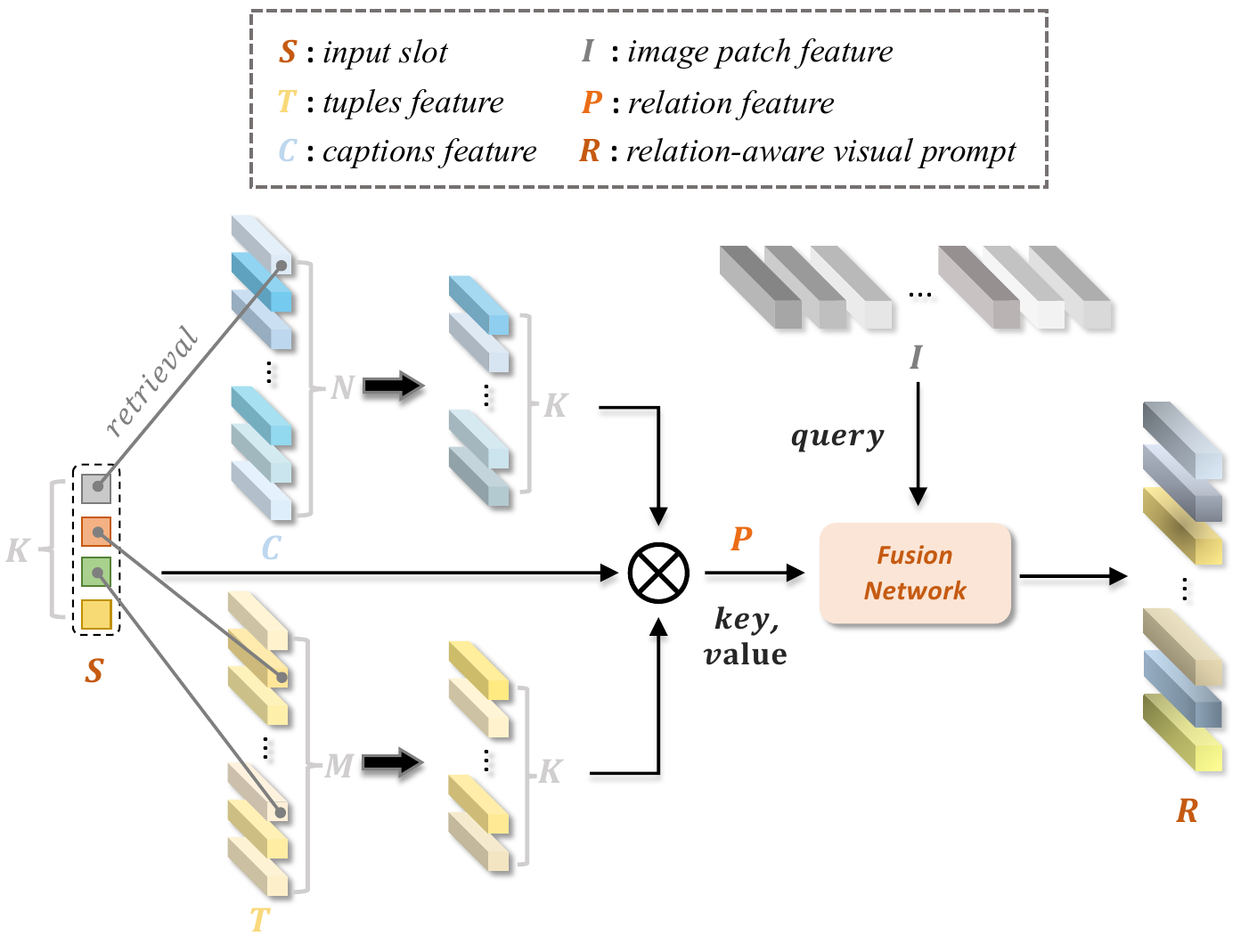}
  \vspace{-0.45cm}
  \caption{The process of slot retrieval and the interaction of image patch feature \textit{I} and relation feature \textit{P} in the Fusion Network.}
\end{figure}

{
\begin{table*}[t]
\centering
\small
\setlength{\tabcolsep}{3pt}
\renewcommand{\arraystretch}{0.92}
\label{tab:main_results}
\makebox[\textwidth][c]{
\begin{tabular}{l|c|cccc|cc|cccc}
\Xhline{1pt} 
\multirow{1.5}{*}{\textbf{Method}} &
\textbf{\shortstack{Training\\Param}} &
\multicolumn{4}{c|}{\textbf{\shortstack{COCO\\Test}}} &
\multicolumn{2}{c|}{\textbf{\shortstack{Flickr30k\\Test}}} &
\multicolumn{4}{c}{\textbf{\shortstack{NoCaps\\Val}}} \\[1pt]
& $\theta$ & BLEU-4 & METEOR & CIDEr & SPICE & CIDEr & SPICE & In-domain & Near-domain & Out-of-domain & Overall \\
\addlinespace[2pt]
\hline
\addlinespace[2pt]
\multicolumn{12}{l}{\textbf{\textit{Heavyweight-training models}}}\\
OSCAR \cite{45_li2020oscar}          & 338M & 37.4 & 30.7 & 127.8 & 23.5 & --   & --   & 78.8 & 78.9 & 77.4 & 78.6 \\
BLIP \cite{17_li2022blip}           & 446M & 40.4 & --   & 136.7 & --   & --   & --   & 114.9 & 112.1 & 115.3 & 113.2 \\
$\text{BLIP2}_\text{$\text{OPT}_\text{2.7B}$}$ \cite{5_li2023blip}   & 1.1B & 43.7 & --   & 145.8 & --   & --   & --   & 123.0 & 117.8 & 123.4 & 119.7 \\
$\text{ViECap}_\text{GPT2}$ \cite{46_fei2023transferable}    & 124M & 27.2 & 24.8 &  92.9 & 18.2 & 47.9 & 13.6 &  61.1 &  64.3 &  65.0 &  66.2 \\
RCA-NOC \cite{47_fan2023rca}       & 110M & 37.4 & 29.6 & 128.4 & 23.1 & --   & --   &  92.2 &  87.8 &  87.5 &  88.3 \\
\addlinespace[2pt]
\hline
\addlinespace[2pt]
\multicolumn{12}{l}{\textbf{\textit{Lightweight-training models}}}\\
ClipCap \cite{7_mokady2021clipcap}        &  43M & 33.5 & 27.5 & 113.1 & 21.1 &  --  &  --  &  84.9 &  66.8 &  49.1 &  65.8 \\
$\text{I-Tuning}_\text{Base}$ \cite{8_luo2023tuning}  &  14M & 34.8 & 28.3 & 116.7 & 21.8 & 61.5 & 16.9 &  83.9 &  70.3 &  48.1 &  67.8 \\
$\text{I-Tuning}_\text{Medium}$ \cite{8_luo2023tuning}  &  44M & 35.5 & \textbf{28.8} & 120.0 & \underline{22.0} & \textbf{72.3} & \textbf{19.0} &  89.6 &  77.4 &  58.8 &  75.4 \\
$\text{SmallCap}_\text{Base}$ \cite{9_ramos2023smallcap}       &   7M & 37.0 & 27.9 & 119.7 & 21.3 & 60.6 &  --  &  87.6 &  78.6 &  68.9 &  77.9 \\
$\text{SmallCap}_\text{Large}$ \cite{9_ramos2023smallcap}       &   47M & 37.2 & 28.3 & 121.8 & 21.5 & -- &  --  &  -- &  -- &  -- &  -- \\
ViPCap \cite{12_kim2025vipcap}        &  14M & \underline{37.7} & \underline{28.6} & \underline{122.9} & 21.9 & 66.8 & 17.2 &  \underline{93.8} &  \underline{81.6} &  \underline{71.5} &  \underline{81.3} \\
\textbf{RACap (Ours)}&  10.8M  &  \textbf{37.9}  &   \textbf{28.8}  &    \textbf{123.0}  &   \textbf{22.1}  &   \underline{69.7}  &   \underline{17.8}  &    \textbf{95.5}  &    \textbf{84.2}  &    \textbf{74.6}  &    \textbf{83.9}  \\
\addlinespace[2pt]
\Xhline{0.9pt}
\end{tabular}}
\vspace{-0.45cm}
\caption{Quantitative comparison with state-of-the-art (SOTA) approaches on the COCO test set, Flickr30k test set, and NoCaps validation set. Higher score is better. Bold indicates the best results, and underlined indicates the second-best results.
}
\end{table*}
}

\begin{figure*}[t]
  \centering
  \includegraphics[width=1\textwidth]{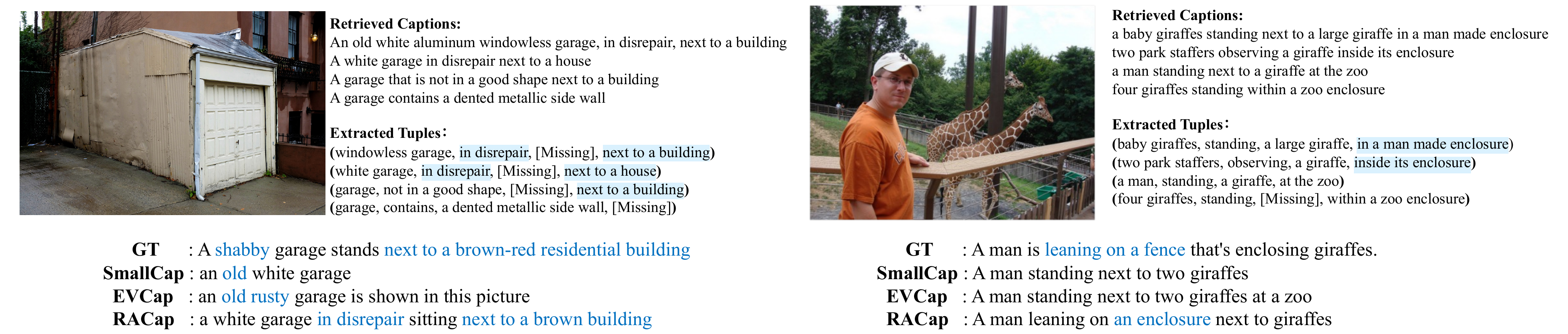} 
  \vspace{-0.6cm}
  \caption{Qualitative Results. RACap can successfully exploit fine-grained objects and their relations in extracted tuples.}
\end{figure*}

\vspace{-0.3cm}
\subsection{Object-Aware Module}
\vspace{-0.35cm}
To identify heterogeneous semantics in the input image, we introduce an object-aware module based on slot attention \cite{13_locatello2020object}. Through multiple iterative steps, a set of learnable slots is progressively updated to capture diverse semantics from the image. The object-aware module uses learnable slot embeddings as queries, while image patch features serve as keys and values\cite{35_kim2023improving}. We apply a total of \textit{2} iterations of attention with shared weights across attention blocks and add a global mean-pooled image feature to each slot at the end of the iterations. In this way, each slot is able to encode distinct object semantics while retaining global contextual information to enable accurate recognition of heterogeneous objects. Define the initial learnable slots as embedding vectors $\boldsymbol{S}^0 \in \mathbb{R}^{K\times D_h}$, where \textit{K} is the cardinality of slots and $D_h$ is the feature dimension. The slots are then updated iteratively for \textit{t}-th steps. The slot attention block linearly projects the input image features \textit{I} into $\boldsymbol{k}\in \mathbb{R}^{N\times D_h}$ and $\boldsymbol{v}\in \mathbb{R}^{N\times D_h}$. The slot embeddings $\boldsymbol{S}^{t-1}$ are linearly projected to $\boldsymbol{q}\in \mathbb{R}^{K\times D_h}$. Specifically, this process can be formulated as:
\begin{equation}
\mathbf{\mathit{Map}}_{n,k} = \frac{e^{\mathit{M}_{n,k}}}{\sum_{i=1}^{K} e^{\mathit{M}_{n,i}}}, \quad \textit{where } \mathit{M} = \frac{\boldsymbol{k} \boldsymbol{q}^{T}}{\sqrt{D_h}}
\label{eq:map_norm}
\end{equation}
\begin{equation}
\mathit{A}_{n,k} = \frac{\mathit{Map}_{n,k}}{\sum_{i=1}^{N} \mathit{Map}_{i,k}}
\label{eq:attn_weights}
\end{equation}

\begin{equation}
\mathit{\boldsymbol{S}}^{t} = \mathit{MLP}\left(\mathit{LN}\left(\mathit{A}^{T} \boldsymbol{v} + \mathit{\boldsymbol{S}}^{t-1}\right)\right) + \mathit{\boldsymbol{S}}^{t-1}
\label{eq:slot_mlp}
\end{equation}
where MLP is a multi-layer perceptron consisting of layer normalization, linear layers, and GELU \cite{49_hendrycks2016gaussian} activation and the subscript denotes the vector shape. Note that the attention map is normalized over the slots, which enables competition of each slot and encourages them to attend to distinct regions of the image. Finally, at the \textit{t} = 2 iteration, the mean-pooled image feature is added to each slot to yield the visual relational representation.

\subsection{S-P-O-E tuple and Slot Retrieval Module}

Unlike previous approaches that focused only on original captions, we leverage a BERT \cite{38_devlin2019bert} model to transform these retrieval captions into S–P–O–E tuples to capture relation information. This strategy enhances both the quality and contextual richness of the retrieved information. Previous methods fail to extract environment part which leads to less complete and semantically accurate captions.
We encode both the retrieval captions and the corresponding S–P–O–E tuples using the CLIP text encoder to form the relation representations. Note that not all captions can be successfully parsed into S–P–O–E tuples, so we insert a \textit{[MISSING]} placeholder to indicate missing components. Although the extracted tuples provide structured relation knowledge, they may not always capture the full richness of the original captions, especially when relation cues of the captions are sparse. Therefore, including the raw caption embedding is essential for preserving complementary information.

To fully exploit the relation information on the captions, we propose a slot retrieval module to select the most relevant semantic features for each slot. In particular, we leverage heterogeneous object slots for independent retrieval on two semantic sources: the retrieval caption features and the S–P–O–E features, as illustrated in Figure 4. 
The slot retrieval module employs cosine similarity to compare the similarity between slot representations \textit{S} and the semantic features \textit{C} and \textit{T}. Given the caption features \textit{C} and S–P–O–E features \textit{T}, this module uses \textit{S} to select \textit{K} related vectors with high similarity from \textit{N} and \textit{M} candidate vectors in \textit{C} and \textit{T} respectively. This strategy not only reduces computational costs but also filters out irrelevant semantics. These selected vectors are then concatenated to form the relational representation P. Note that we also concatenate the original slot features. This residual information helps preserve object-level visual features and maintains visual modality information during the subsequent fusion network.

\subsection{Fusion Network and Model Training}
We use a Fusion Network (FN) to fuse the relational representation \textit{P} and further reduce the noises in it. FN is designed as a single-layer multi-head cross-attention module followed by an MLP. The image features \textit{I} serve as the query, while the relational features \textit{P} are used as the key and value. The output is further processed by an MLP to obtain the final visual prompt \textit{R}.
The hard prompt is organized as follows: \textit{``Similar images show ${caption_1}...{caption_k}$. This image shows \_\_\_''}. During training, the decoder minimizes the cross-entropy loss over the \textit{N} predicted reference tokens $y_{1}, \ldots, y_{N}$ to optimize the weights of the cross-attention layer ($\theta$):

\begin{equation}
\mathcal{L}_\theta = -\sum_{i=1}^{N} \log P_\theta(y_i|y_{<i}, R; \theta)
\label{eq:loss}
\end{equation}

\begin{table}[t]
\centering
\small
\label{tab:ablation_all}
\setlength{\tabcolsep}{3pt}
\renewcommand{\arraystretch}{0.9}
\begin{tabular}{l|cc|cc}
\Xhline{1pt}
\addlinespace[2pt]
\multirow{2}{*}{\textbf{Method}} &
\multicolumn{2}{c|}{\textbf{COCO Test}} &
\multicolumn{2}{c}{\textbf{Flickr30k Test}} \\
 & \text{BLEU@4} & \text{CIDEr} & \text{CIDEr} & \text{SPICE} \\
\addlinespace[2pt]
\Xhline{0.8pt}
\addlinespace[2pt]
\multicolumn{5}{l}{\textbf{Implementation of object-aware module}} \\
Raw learnable Slots    & 37.2 & 120.5 & 61.2 & 16.6 \\
Multi-head attention   & 37.3 & 120.9 & 62.3 & 16.7 \\
\addlinespace[4pt]
\hline
\addlinespace[2pt]
\multicolumn{5}{l}{\textbf{Extracted textual information}} \\
Object name        & 37.2 & 120.2 & 68.2 & 17.1 \\
S-P-O Triplets     & 37.4 & 121.5 & 68.4 & 17.5 \\
\addlinespace[4pt]
\hline
\addlinespace[2pt]
\multicolumn{5}{l}{\textbf{The retrieval scope of slot retrieval module}} \\
w/o Retrieval    & 37.2 & 120.9 & 65.1 & 16.7 \\
Retrieval Cap Only       & 37.4 & 121.2 & 66.6 & 16.7 \\
Retrieval Tuple Only    & 37.4 & 121.7 & 68.8 & 17.3 \\
\addlinespace[4pt]
\hline
\addlinespace[2pt]
\multicolumn{5}{l}{\textbf{Fusion strategies in FN}} \\
Sum the features      & 36.8 & 120.2 & 63.1 & 16.9 \\
Concat+MLP & 37.0 & 121.3 & 65.6 & 17.0 \\
Multi-head attention        & 37.4 & 122.5 & 68.1 & 17.4 \\
\addlinespace[4pt]
\hline
\addlinespace[2pt]
\textbf{RACap(Ours)}      & \textbf{37.9} & \textbf{123.0} & \textbf{69.7} & \textbf{17.8} \\
\addlinespace[2pt]
\Xhline{1pt}
\end{tabular}
\vspace{-0.25cm}
\caption{Comprehensive ablation study on four components.}
\end{table}

\section{Experiments}
\label{sec:typestyle}

\subsection{Dataset and Setup}
We conduct experiments on three benchmark datasets: COCO \cite{14_lin2014microsoft}, Flickr30k \cite{15_plummer2015flickr30k}, and NoCaps \cite{16_agrawal2019nocaps}. Following \cite{9_ramos2023smallcap,11_li2024evcap,12_kim2025vipcap}, we train RACap using only the training split of the COCO dataset, and evaluate the model on the  NoCaps validation set, COCO test set, and Flickr30k test set using the Karpathy split \cite{37_karpathy2015deep}.
RACap uses frozen CLIP-ViT-B/32 as the image encoder and $\text{GPT2}_\text{base}$ for text generation and only train the object-aware module with its slots, FN, and the cross-attention layer. Following the
configuration in SmallCap, the cross-attention layer consists of a 12-head layer, and the hidden
dimension is reduced to 16. The entire model is trained on a NVIDIA RTX 4090 GPU. We use the AdamW optimizer \cite{39_loshchilov2017decoupled} for a total of 10 epochs with a batch size of 80.
During retrieval, RACap uses CLIP-ResNet-50x64 to retrieve 4 captions for each image. FAISS \cite{40_johnson2019billion} is used for efficient nearest neighbor searching. During inference, RACap uses beam search decoding with a beam size of 3 to generate captions. Standard metrics including BLEU@4  \cite{41_papineni2002bleu}, METEOR (M) \cite{42_denkowski2014meteor}, CIDEr (C) \cite{43_vedantam2015cider} , and SPICE (S) \cite{44_anderson2016spice} are adopted.

\subsection{Quantitative Results}
We evaluate the in-domain performance of RACap on the COCO test set and the out-of-domain results on the Flickr30k test set and the NoCaps validation set. As shown in Table 1, RACap achieves superior performance despite having significantly fewer parameters. Compared to SmallCap, RACap not only outperforms $\text{SmallCap}_\text{Base}$ but also surpasses $\text{SmallCap}_\text{Large}$, which has substantially larger parameters. Compared to $\text{I-Tuning}_\text{base}$ with 14M parameters, our model exhibits better performance with 3M fewer trainable parameters. Notably, on the NoCaps dataset, RACap outperforms a range of state-of-the-art lightweight models, exceeding the second-best method by over 2 points. 

Moreover, when compared with heavyweight-training models, RACap remains competitive with only 10.8M trainable parameters. For instance, RACap achieves comparable performance to $\text{OSCAR}_\text{Large}$, with a CIDEr gain of over 16 points on in-domain part in NoCaps set, while using approximately 1/30 of its trainable parameters. These results highlight that despite its smaller training cost, RACap exhibits strong and competitive efficiency and transfer capabilities.

\subsection{Qualitative Results}
Figure 4 presents the image captioning examples of RACap, SmallCap, and EVCap for images in the COCO test set and NoCaps validation set. Due to the extraction of tuples, compared with SmallCap which only retrieves information from the original captions and EVCap which recognizes the object name, RACap can recognize fine-grained objects more clearly. For example, \textit{building} in the first image and \textit{enclosure} in the second image. Moreover, in addition to fine-grained recognition, RACap can successfully describe relationships between multiple objects. These two examples clearly illustrate the effectiveness of RACap.

\subsection{Ablation Studies}
To evaluate the effectiveness of the components proposed in our model, we conducted a series of ablation studies as shown in Table 2. Firstly, we analyze the effect of the object-aware module by changing the different processing strategies of learnable embeddings: raw learnable embeddings (no processing) and multi-head attention over embeddings. Secondly, we extract different levels of semantic structure: extract object names and extract Subject-Predicate-Object (S-P-O) triplets. What's more, we switch different retrieval configurations in the slot retrieval module: no retrieval, retrieval of captions only, and retrieval of tuples only. Finally, we explore the effect of fusion strategies in the Fusion Network, including summation of features, concatenation with MLP, and attention-based fusion only. Results in Table 2 demonstrate the superiority of the design of RACap including the object-aware module, S-P-O-E tuples, the slot retrieval module, and the design of FN.

\section{Conclusion}
\label{sec:majhead}

In this work, we propose RACap, a novel image captioning method that extracts relational information from retrieval captions and image features. Our proposed object-aware module captures heterogeneous semantics from image features, which are then combined with the S-P-O-E and caption semantics. These features are fed into the slot retrieval module to perform cross-modal retrieval between visual and textual presentations. Experimental results show that RACap achieves strong in-domain and out-of-domain performance.

\section{Acknowledgements}
The work was supported by ``Tianshan Talent'' Training Program (2023TSYCLJ0030).

\vfill\pagebreak

\bibliographystyle{IEEEbib}
\bibliography{refs}

\end{document}